\documentclass[10pt,twocolumn,letterpaper]{article}

\usepackage[pagenumbers]{cvpr} 

\usepackage{graphicx}
\usepackage{amsmath}
\usepackage{amssymb}
\usepackage{booktabs}
\usepackage{wrapfig}

\usepackage{stfloats}
\usepackage{url}            
\usepackage{tabularx}
\usepackage{amsfonts}       
\usepackage{nicefrac}       
\usepackage{microtype}      
\usepackage{xcolor}         
\usepackage{amsmath,amsfonts,amsthm,color}
\usepackage[toc]{appendix}
\newcommand{\ra}[1]{\renewcommand{\arraystretch}{#1}}

\definecolor{cvprblue}{rgb}{0.21,0.49,0.74}
\usepackage[pagebackref,breaklinks,colorlinks,allcolors=cvprblue]{hyperref}

\usepackage[capitalize]{cleveref}
\crefname{section}{Sec.}{Secs.}
\Crefname{section}{Section}{Sections}
\Crefname{table}{Table}{Tables}
\crefname{table}{Tab.}{Tabs.}

\def\paperID{*****} 
\def\confName{CVPR}
\def\confYear{2025}

\title{Scalable and Realistic Virtual Try-on Application for \\
Foundation Makeup with Kubelka-Munk Theory}

\author{Hui Pang
\and
Sunil Hadap
\and
Violetta Shevchenko
\thanks{Work done at Amazon}
\and
Rahul Suresh
\footnotemark[1]
\and
Amin Banitalebi-Dehkordi
\\
\\
Amazon
}

\begin{document}
\maketitle

\begin{abstract}
    Augmented reality is revolutionizing beauty industry with virtual try-on (VTO) applications, which empowers users to try a wide variety of products using their phones without the hassle of physically putting on real products. A critical technical challenge in foundation VTO applications is the accurate synthesis of foundation-skin tone color blending while maintaining the scalability of the method across diverse product ranges. In this work, we propose a novel method to approximate well-established Kubelka-Munk (KM) theory for faster image synthesis while preserving foundation-skin tone color blending realism. Additionally, we build a scalable end-to-end framework for realistic foundation makeup VTO solely depending on the product information available on e-commerce sites. We validate our method using real-world makeup images, demonstrating that our framework outperforms other techniques.  

\end{abstract}

\section{Introduction}
\label{sec:intro}

\begin{figure}[ht!]
  \centering
  \includegraphics[width=\linewidth]{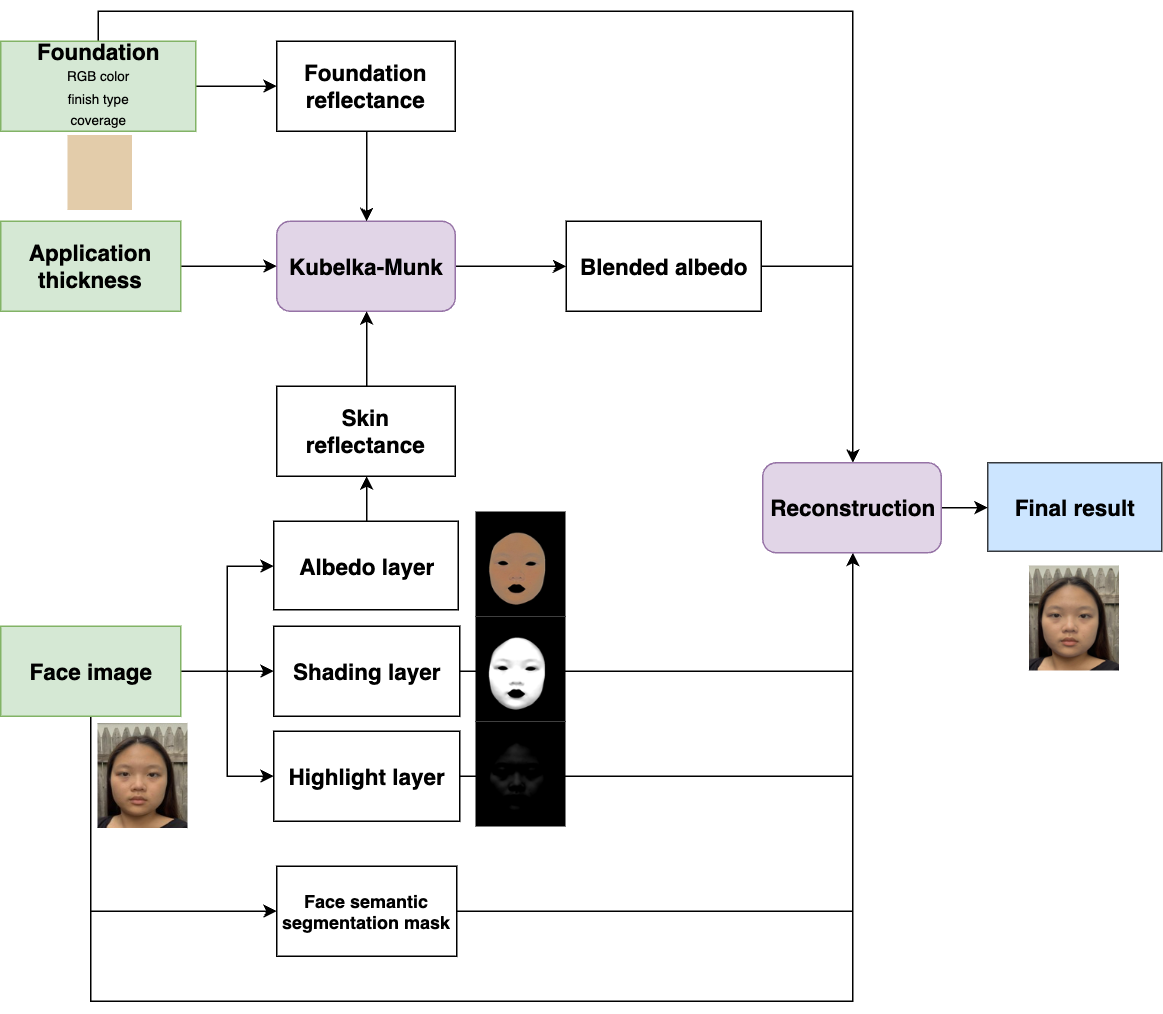}
  \caption{End-to-end framework of foundation virtual try-on application. First, we create a face semantic segmentation mask and decompose face image into albedo, shading and specular highlight layers. Then we estimate the reflectance of albedo layer as well as the reflectance of foundation product based on sRGB values. Then we utilize our approximation for Kubelka-Munk model to blend foundation shade with albedo. Finally, we reconstruct the image with blended albedo, shading and specular highlight.}
  \label{fig:e2e_framework}
\end{figure}

Virtual try-on (VTO) applications have transformed the way users shop for beauty products online. Prior to the advent of VTO, shoppers were limited to relying on static product images and reviews, often resulting in uncertainty about color compatibility and overall look. VTO applications provide a realistic simulation of how the product would appear on face, enabling informed buying decisions. 

Color accuracy is crucial to foundation virtual try-on applications. Although recent makeup transfer models, e.g. BeautyREC~\cite{yan2022beautyrecrobustefficientcontentpreserving}, EleGANT~\cite{yang2022elegant} and SSAT~\cite{sun2021ssatsymmetricsemanticawaretransformer} demonstrate proficiency in color and pattern transfer, they fail to account for the critical interaction between underlying skin tones and foundation products (See \cref{fig:results}). This limitation significantly impacts their reliability as decision-making tools for consumer purchases. 
To extract accurate skin tones from images, there have been multiple works trying to disentangle skin albedo color from shading and highlight in an image \cite{dasPIENet,Li14, NeuralFace2017}. A significant challenge persists in the accurate representation of various skin tones. Beside accurate skin tone extraction, VTO applications must accurately show the color of foundation applied on skin. While alpha blending\cite{Wallace81} has been widely adopted for its computational efficiency and implementation simplicity, it fails to capture the complex optical interactions of real-world foundation shade blending with skin tone. At the same time, the acquisition of optical properties for cosmetic products are expensive, as the measurement process requires specialized equipment and extensive laboratory analysis~\cite{Curtis97, Baxter04, Doi05, Huang13, Lu14, Tan19}. Therefore, another challenge in constructing VTO applications lie in striking a balance between delivering highly realistic results and ensuring the technology is scalable to accommodate diverse beauty products and skin tones. 

In this paper, we propose an end-to-end framework that achieves realistic rendering of foundation products relying only on the product information available on e-commerce sites. We first trained a face semantic segmentation model to locate the region for the product application, and developed a simple and fast intrinsic image decomposition method to separate the face skin tone from the shading and specular highlight effects. (\cref{subsec: face semantic segmentation and intrinsic image decomposition}). Second, we utilized Kubelka-Munk (KM)~\cite{KM31} theory for realistic color blending of foundation shade and skin tone (\cref{subsec: color blending}). Since KM theory is formulated in terms of reflectance spectra and optical properties of blended materials, we developed a method to estimate the reflectance of skin and makeup directly from their sRGB values (\cref{subsec: reflectance-simplification}) and to estimate foundation scattering coefficient based on its coverage level (\cref{subsec: scattering-coefficient-estimation}). Finally, to reduce the computational complexity of the proposed approach, we devised a novel method to approximate the KM model with Taylor Expansion (\cref{subsec: taylor-expansion-approximation}). 

In summary, our contributions can be summarized as:
\begin{itemize}
    \item We propose a scalable end-to-end framework that achieves realistic virtual try-on experience for foundation makeup (\cref{sec: e2e framework});
    \item We develop a method to approximate Kubelka-Munk model with Taylor Expansion which significantly reduces the computation complexity while preserving the realism of color blending (\cref{subsec: taylor-expansion-approximation}).
\end{itemize}

\section{Related Work}
Multiple works have explored virtual try-on for makeup, broadly categorized as learning-based makeup transfer neural networks \cite{liu2021psgan++, Zhang2019DisentangledMT, nguyen2021lipstickaintenoughcolor} and physics-based methods e.g., using Kubelka-Munk theory \cite{Doi05, Huang13}. Compared to learning-based models, physics-based methods provide more robust synthesis results, and Kubelka-Munk theory plays a critical role in ensuring realistic color of makeup blended with skin. 

Kubelka and Munk's seminal work~\cite{KM31} introduced a groundbreaking approach for realistically simulating the reflectance of a paint applied on a substrate. However, widespread adoption of Kubelka-Munk model has been hindered by its implementation complexity. Firstly, it requires multispectral measurements of reflectances in a controlled environment, which can be expensive and time-consuming. Secondly, reflectance and transmittance depend on absorption and scattering coefficients, and computing and storing the absorption and scattering coefficients for each wavelength throughout the spectrum can be extremely expensive.

To address this problem, several works proposed to estimate reflectance directly from sRGB values. Such methods can be divided into optimization and data-driven approaches. Optimization approaches try to find a spectrum that minimizes a cost function. Smits~\cite{Smits1999AnRC} approached this with a linear optimization of discretized spectra based on the observation that natural spectra are often smooth~\cite{Maloney86}. Dupont~\cite{Dupont02} evaluated multiple optimization methods and found Hawkyard \cite{Hawkyard2008SyntheticRC} was among the best. Data-driven approaches~\cite{Cohen64,Otsu18} require pre-measured spectra data and propose to represent reflectance as a weighted sum of basis functions.

Regarding the absorption and scattering coefficient computation, there are mainly two streams of research. The first one requires experimenting with pigments on different backgrounds. Curtis et al.~\cite{Curtis97} applied pigments over a black and a white background, then estimated the absorption and scattering coefficients directly with the sRGB colors. \cite{Doi05, Huang13} measured the reflectance spectrum directly with spectrophotometer/spectroradiometer and derived the optical coefficients by comparing the reflectance of a thick layer and a thin layer over a black background. The second line of research relies on the idea of mixing primary pigments. Baxter et al.~\cite{Baxter04} decomposed an arbitrary oil paint as a mixture of the 11 standard oil paints, which reflectances where pre-measured. Lu et al.~\cite{Lu14} used three distinct primary pigments and approximated KM model with a data-driven method using a physical color chart. Tan et al.~\cite{Tan19} discovered the primary pigments in a painting by finding 3D convex hull of image sRGB colors, then estimated the mixing weights of an arbitrary color. Šárka Sochorová and Ondřej Jamriška \cite{Soch21} defined a color as a linear combination of primary sRGB colors which absorption and scattering coefficients are known, then composited two colors by interpolating the weights of primary colors.

The aforementioned approaches all rely on supplementary information, such as direct optical measurements requiring specialized equipment, or pre-existing optical data on pigments. Such methods are not scalable for real-world applications and do not fit the online e-commerce VTO use case. In this paper, we propose a method that does not rely on any auxiliary data and can achieve realistic color blending results by using only the product information available on e-commerce sites. 

\section{End-to-end Framework for Foundation Virtual Try-on Application}
\label{sec: e2e framework}
There are two key challenges to achieve a realistic virtual try-on experience: 1) maintaining consistent light distribution across the user's face post-application and ensuring the preservation of facial details like eyebrows and eyelashes; 2) achieving accurate color blending of foundation shade with the user's original skin tone. In this section, we describe our end-to-end framework (\cref{fig:e2e_framework}) which includes face semantic segmentation, intrinsic image decomposition and color blending.

\begin{figure*}[ht!]
  \centering
  \includegraphics[width=0.70\linewidth]{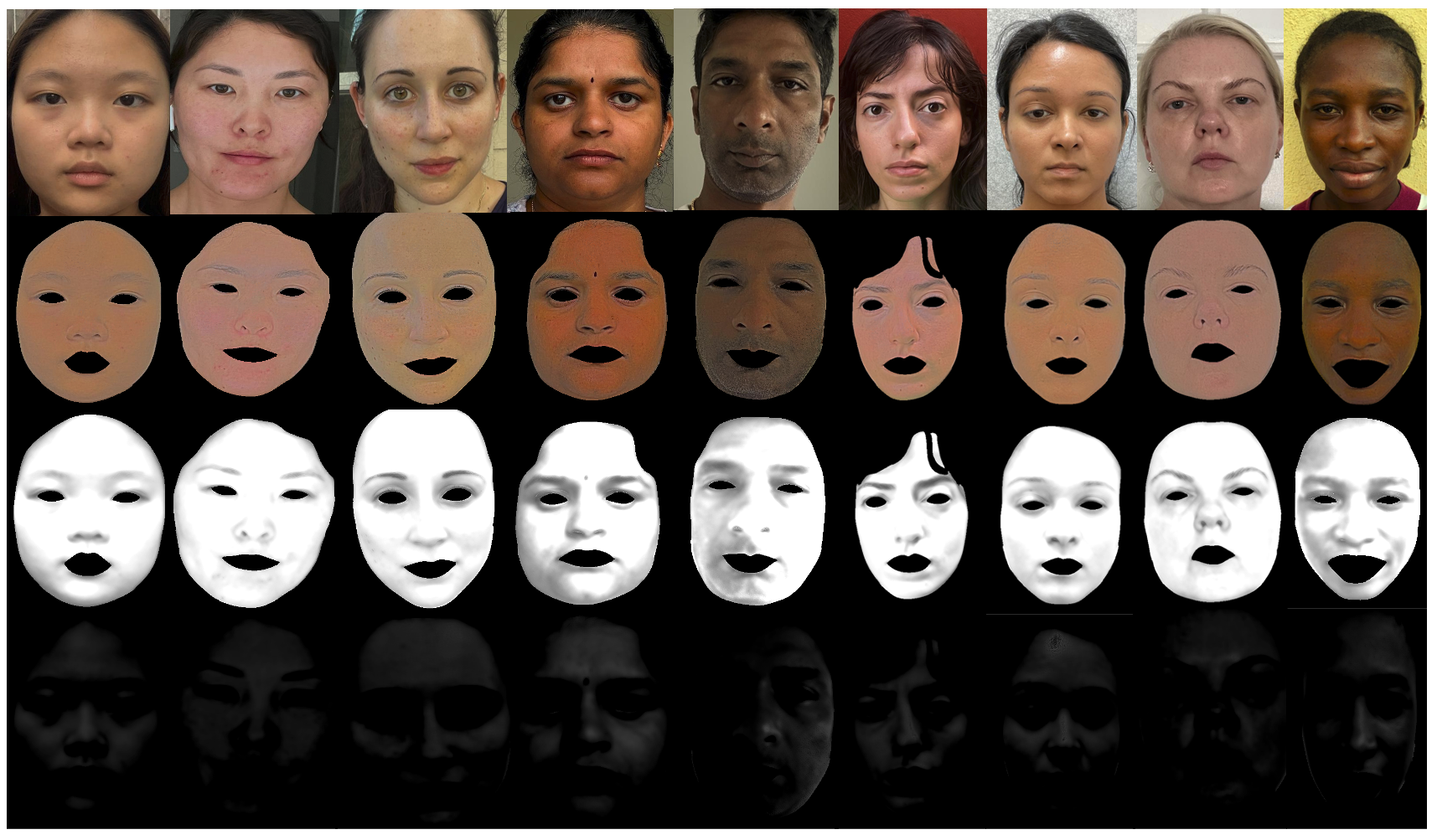}
  \caption{Examples of face image intrinsic decomposition: First row is original face image; second row is albedo layer; third row is shading and forth row is specular highlight.}
  \label{fig:image_decomposition_results}
\end{figure*}

\subsection{Face semantic segmentation and intrinsic image decomposition}
\label{subsec: face semantic segmentation and intrinsic image decomposition}
To train a face semantic segmentation model, we collected a dataset of 4897 face images. Annotators then labeled the images with semantic segmentation masks, assigning 17 distinct labels corresponding to various facial regions. We compared several baselines (BiSeNetV2~\cite{yu2021bisenet}, U-Net~\cite{ronneberger2015u}, DeepLabv3~\cite{chen2017rethinking}, FaRL~\cite{zheng2022general}), and selected BiSeNetV2 model for our pipeline as it obtained the best balance between model size and segmentation accuracy. Please see \cref{fig:semantic-segmentation} and \cref{tab:face_segmentation} in supplementary materials for details.

Intrinsic image decomposition aims to extract the skin tone (albedo), shading and specular highlight from a face image. This step is required to ensure accurate color blending using the original skin tone that is not affected by lighting conditions. Following~\cite{Li15}, we formulate the decomposition of an image into three intrinsic layers as: 

\begin{equation}\label{eq:image_decompostion}
    I = A*S + H,
\end{equation}
where \(I\) is the original image, \(A\) is albedo, \(S\) is shading and \(H\) is specular highlight.

Image decomposition is a severely under-constrained problem as it lacks a unique solution. There are multiple combinations of albedo, highlights and shading that can explain the observed image equally well. However, we can explicitly enforce some constraints and obtain an approximate solution which proved to be sufficient in our case.
First, we utilize image processing method\cite{ShadowHighlightCorrection} to remove shading and highlight from the face to get an estimate of albedo layer \(A'\). We determine the amount of shading and highlight to be removed by the median value of the grayscale image \(s^*=Median(I_{gray})\), i.e. the region darker than \(s^*\) is taken as shading and the region brighter than \(1 - s^*\) is taken as highlight. Second, for each pixel, we solve the equations below to compute the approximate shading \(S'\) and specular highlight \(H'\) using the least square optimization with constraints:

\begin{equation} \label{eq:image_decomposition_ls}
    \begin{bmatrix} 1 & r_{albedo}\\ 1 & g_{albedo} \\ 1 & b_{albedo} \end{bmatrix} \begin{bmatrix} x_{specular}\\ x_{shading}\end{bmatrix} = \begin{bmatrix} r_{image} \\ g_{image} \\ b_{image}\end{bmatrix}
\end{equation}
where \(0 \leq x_{shading} \leq 1\) and \(0 \leq x_{specular} \leq 1\).

We made an assumption that both shading and highlight are grayscale, therefore, some color information can be lost during reconstruction. To resolve this problem, we recover the lost information and compute the new highlight layer as \(H' = I - A'*S'\), following \cref{eq:image_decompostion}. However, when the highlights are too strong and appear as purely white pixels in the image, the information about the original albedo color becomes irretrievably lost. For that reason, we adopted LaMa inpainting model~\cite{suvorov2021resolution} to fill in the missing skin color based on the surrounding pixels not affected by highlights. Specifically, we mask the highlighted areas in the image and pass it through the inpainting model to recover the colors (see \cref{fig:image_decomposition_results} for intrinsic image decomposition examples). As the inpainting model's performance depends on the highlight mask, we use the predicted highlight layer $H'$ as a guidance for precise mask generation. As we preserve the shading area while removing the highlight, we don't need to recompute the shading layer (\cref{eq:image_decomposition_ls}) after using inpainting model.


\subsection{Color blending}
\label{subsec: color blending}
Following the intrinsic decomposition of a facial image we leverage the albedo layer, which represents the base skin color, to apply the foundation color. To blend the foundation and skin color, we utilize the novel approach presented in \cref{sec: KM for VTO}. Next, combining the blended foundation albedo layer with the original shading and highlight layers, we reconstruct the full face image as \(I_{afterMakeup} = A_{afterMakeup} * S' + H'\). To create looks for foundation with different finish types, e.g. matte and glowy, we adjust the intensity of specular highlight layer $H'$ to control the amount of reflection on the face. Finally, to ensure that foundation color is applied on skin areas only, we composite the resulting image with the face semantic segmentation mask: \(I_{result} = Mask * I_{afterMakeup} + (1-Mask) * I\).

\section{Adapting Kubelka-Munk Theory for Virtual Try-On}
\label{sec: KM for VTO}
\begin{figure}[ht!]
  \centering
  \includegraphics[width=1\linewidth]{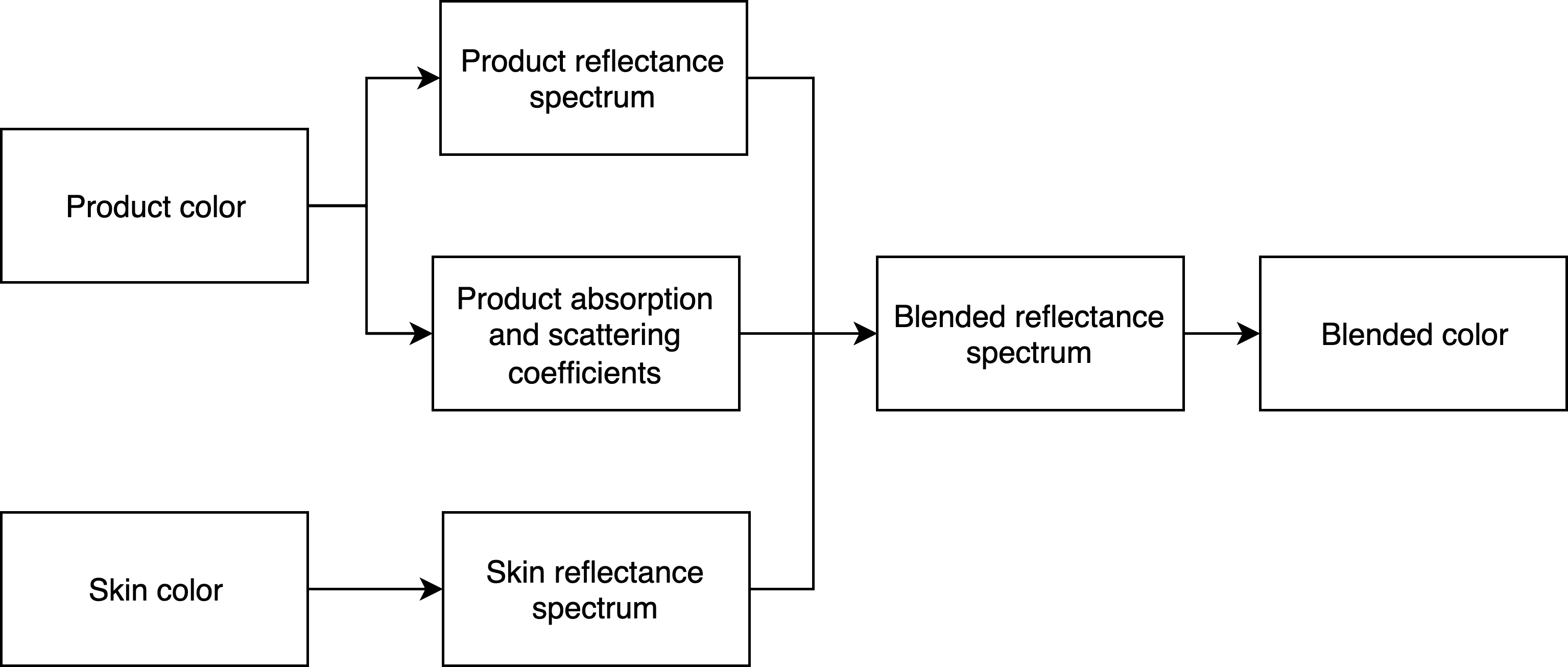}
  \caption{Workflow of conventional implementation of KM model.}
  \label{fig:KM_workflow}
\end{figure}

In this section, we adapt the Kubelka-Munk (KM) theory for realistic color blending of foundation and user's skin tone. To use KM theory (See \cref{fig:KM_workflow}), we need to estimate the reflectance of both skin and foundation, as well as transmittance of foundation which is determined by its scattering coefficient when the color is fixed. Firstly, we present a novel approach to estimate the reflectance spectrum using only readily available sRGB color values (\cref{subsec: reflectance-simplification}). Secondly, we describe a new method to estimate scattering coefficient based on the coverage property of foundation makeup (\cref{subsec: scattering-coefficient-estimation}). Finally, we introduce an innovative method that leverages Taylor Expansion to significantly improve the computational efficiency of Kubelka-Munk model (\cref{subsec: taylor-expansion-approximation}).

\subsection{Kubelka-Munk theory preliminary}
Based on Kubelka-Munk theory \cite{KM31}, the reflectance of an infinitely thick (opaque) layer is
\begin{equation}\label{eq:reflectance_infinity}
R_{\infty}(\lambda) = 1+\frac{K(\lambda)}{S(\lambda)} - \sqrt{\frac{K(\lambda)^2}{S(\lambda)^2}+\frac{2K(\lambda)}{S(\lambda)}},
\end{equation}
where \(S\) is the wave-length dependent scattering coefficient and \(K\) is the wave-length dependent absorption coefficient.

For a layer of finite thickness, according to \cite{Gustav69}, the reflectance and transmittance of thickness $D$ can be measured as:

\begin{equation}\label{eq:reflectance_x}
R(\lambda)=\frac{sinh\ b(\lambda)S(\lambda)D}{a(\lambda)sinh\ b(\lambda)S(\lambda)D + b(\lambda)cosh\ b(\lambda)S(\lambda)D}
\end{equation}

\begin{equation}\label{eq:transmittance_x}
T(\lambda) = \frac{b(\lambda)}{a(\lambda)sinh\ b(\lambda)S(\lambda)D + b(\lambda)cosh\ b(\lambda)S(\lambda)D}
\end{equation}

where \(a = 1+ \frac{K}{S}\) and \(b = \sqrt{a^2 - 1}\).

\begin{figure}[ht]
  \centering
  \includegraphics[width=0.85\linewidth]{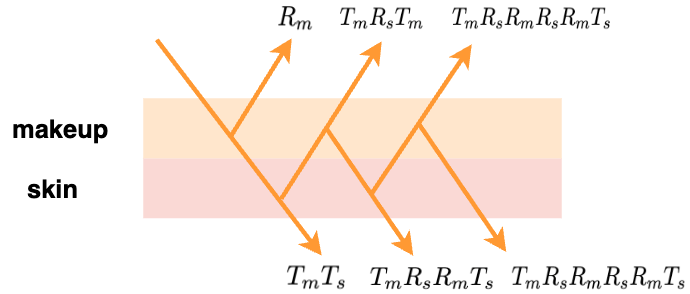}
  \caption{An illustration of the reflectance and transmittance of the foundation layer applied on skin.}
  \label{fig:skin_foundation}
\end{figure}

According to ~\cite{Kubelka54}, the blended reflectance $R$ of two non-homogeneous layers (skin and foundation in our use case) is defined as (see \cref{fig:skin_foundation} for a visualization):

\begin{align} \label{eq:km_blending}
    R & = R_m + T_mR_sT_m+T_mR_sR_mR_sT_m+...... \\
      & = R_m + \frac{T_m^2R_s}{1-R_mR_s},              \label{eq:km_blending2}
\end{align}
where $R_s$ is the reflectance of skin, $R_m$ and $T_m$ are reflectance and transmittance of the foundation foundation respectively. 

\begin{figure}[ht!]
  \centering
  \includegraphics[width=170pt]{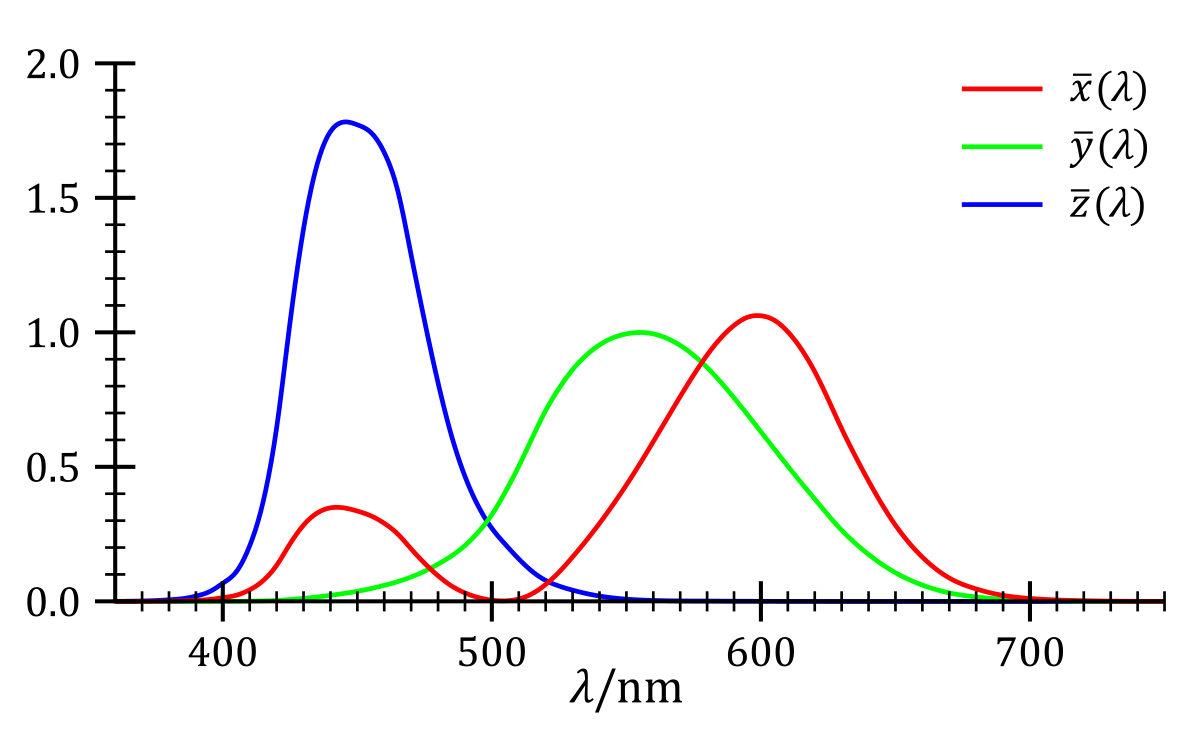}
  \caption{The CIE XYZ standard observer color matching functions \cite{ac90}.}
  \label{fig:eye_sensitivity}
\end{figure}

Because the color we perceive depends on the lighting of the environment, the reflectance of an object and the sensitivity of human eyes to different wavelengths of light. \cref{eq:reflectance_xyz_conversion} shows a discrete version of how to compute color from reflectance given $D_{65}$ standard daylight illuminant \cite{Soch21}.

\begin{equation}\label{eq:reflectance_xyz_conversion}
\left \{ 
\begin{array} {l} 
\sum_{\lambda}D_{65}(\lambda)  \bar x(\lambda) R (\lambda) = X \\ 
\sum_{\lambda} D_{65}(\lambda) \bar y (\lambda)  R (\lambda) = Y  \\  
\sum_{\lambda} D_{65}(\lambda)\bar z(\lambda) R (\lambda) = Z
\end{array} 
\right.
\end{equation}
where \(D_{65}(\lambda)\) is the standard daylight illuminant, \(\bar{x}(\lambda)\), \(\bar{y}(\lambda)\) and \(\bar{z}(\lambda)\) are CIE XYZ standard observer color matching functions \cite{ac90} (\cref{fig:eye_sensitivity}), and \(R(\lambda)\) is the reflectance spectrum of an object.

\subsection{Reflectance estimation}
\label{subsec: reflectance-simplification}

Kubelka-Munk theory relies on object reflectance information. To leverage this theory, we need to convert sRGB color to a reflectance spectrum. To simplify the problem, we propose to use piece-wise linear function to approximate the reflectance.

\begin{figure}[ht!]
  \centering
  \includegraphics[width=0.85\linewidth]{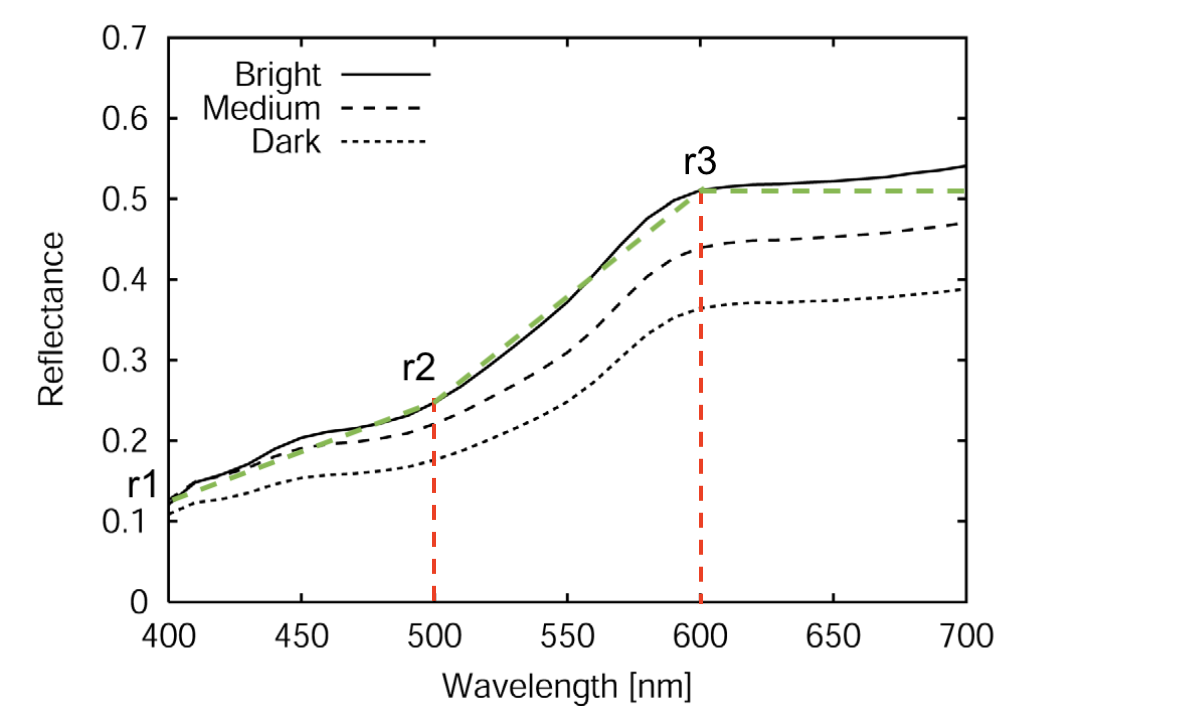}
  \caption{The black lines are the reflectance examples of bright, medium and dark foundation from Doi et al. \cite{Doi05}; the green dotted line represents our approximation of the reflectance spectrum.}
  \label{fig:foundation_reflectance}
\end{figure}


As shown in \cref{fig:foundation_reflectance}, we split the whole spectrum into pieces, and for each piece we use a linear function. We observed that foundation reflectance and most skin reflectances \cite{Cooksey17} remain relatively constant for longer wavelengths, we thus set the third piece to be flat, meaning that we only need three parameters ($r_1, r_2, r_3$) to estimate the full visible spectrum:

\begin{equation}
\label{eq:piecewise_linear_approximation}
R(\lambda) = \begin{cases}  
\frac{r_2 - r_1}{100}(\lambda - 400) + r_1, & 400 \leq \lambda < 500 \\  
\frac{r_3-r_2}{100}(\lambda - 500)+ r_2, &  500 \leq \lambda < 600 \\  
r_3, &  600 \leq \lambda \leq 700  
\end{cases},
\end{equation}
where \(\lambda\) is the wavelength.

We solve ($r_1, r_2, r_3$) based on the relationship between reflectance spectrum and perceived colors. We first convert sRGB to XYZ color space. Using our piece-wise linear approximation for reflectance spectrum, from \cref{eq:reflectance_xyz_conversion} and \cref{eq:piecewise_linear_approximation} , we can formulate:

\begin{equation}\label{eq:piecewise_linear_approximation_matrix}
    \begin{bmatrix} 
        X \\ Y \\ Z 
    \end{bmatrix}  
    =  
    \begin{bmatrix}  
        A & B & C  
    \end{bmatrix} 
    \begin{bmatrix} 
        r_1 \\ r_2 \\ r_3
    \end{bmatrix},
\end{equation}
where \(A\), \(B\) and \(C\) can be computed through known constants and we only need to compute them once. Then we can solve the equations to obtain \(r_1\), \(r_2\) and \(r_3\).

\subsection{Scattering coefficient estimation}
\label{subsec: scattering-coefficient-estimation}
Assuming skin is opaque, skin reflectance spectrum can be estimated directly from sRGB values in the image using the method described in \cref{subsec: reflectance-simplification}. However, foundation makeup applied on skin is a thin layer whose color is different from the color extracted from product image, and its reflectance and transmittance depend on foundation properties and the thickness of the applied layer. Thus we will derive the reflectance and transmittance of a finite layer of foundation based on its product color extracted from product image which we assume corresponds to a fully opaque layer of foundation.

Assuming that \(R_{\infty}(\lambda)\) corresponds to the sRGB color we extracted from the foundation product image because the product is opaque, we can estimate it using the method described in \cref{subsec: reflectance-simplification}, thus compute \(\frac{K(\lambda)}{S(\lambda)}\) ratio from \cref{eq:reflectance_infinity}, and subsequently, compute $a$ and $b$ values. Given the pre-defined thickness $D$ and foundation color of an opaque layer, the only unknown in \cref{eq:reflectance_x} and \cref{eq:transmittance_x} is the scattering coefficient $S$ and it defines how much skin we can see through the foundation based on \cref{eq:km_blending2}. Besides sRGB color, there is another inherent property of foundation -- coverage, which determines the opacity of the foundation. Higher coverage foundation products are more opaque, covering more skin color. Therefore, we can hypothesize that the scattering coefficient \(S\) is correlated with the foundation coverage property.

We assume that for the same thickness and coverage, the reflectance of different foundation colors will have the same proportion of its corresponding reflectance of infinite thickness \(R_{\infty}\). Additionally, for the same thickness, the reflectance of high coverage foundation compared to low coverage foundation should have a higher proportion of the reflectance of infinite thickness, since high coverage foundation is more opaque. Empirically, for fixed $D=0.25$, we set the reflectance $R$ of full coverage foundation to be 30\% of $R_{\infty}$, 20\% for medium coverage and 10\% for low coverage. Then we can derive the scattering coefficient with a fixed thickness as \(S = \frac{coth^{-1}(\frac{1-aR}{bR})}{bD}\).

\subsection{Approximation of Kubelka-Munk model}
\label{subsec: taylor-expansion-approximation}

 The major drawback of the KM implementation is the expensive computation of the per-wavelength variables and the necessity to integrate over the full visible spectrum to obtain the final blended color. To mitigate this problem, we propose a novel method to approximate \cref{eq:km_blending2} by Taylor Expansion. Our approximation allows to skip the cumbersome integration over the reflectance spectrum and to directly make use of XYZ color channels of the skin and foundation makeup, which significantly simplifies the computation. 

Rewriting \cref{eq:km_blending} gives: 
\begin{equation}
R = R_m + \frac{T_m^2 R_s}{1-R_mR_s}=R_m+f(R_m, R_s),
\end{equation}
where $f(R_m, R_s) = \frac{tR_s}{1 - R_mR_s}$ and $t = T_m^2$.

Based on Taylor Expansion,
\begin{align}\label{eq:te}
    \begin{split}
        & \quad f(R_m, R_s) \\
        & = f(r_m, r_s) + f_{R_m}(r_m, r_s)(R_m-r_m) \\
        & \quad +f_{R_s}(r_m, r_s)(R_s - r_s)+\epsilon \\
        & = f_{R_m}(r_m, r_s)R_m + f_{R_s}(r_m, r_s)R_s \\
        & \quad + \big( f(r_m,r_s) - r_mf_{R_m}(r_m, r_s)-r_s f_{R_s}(r_m, r_s) \big) \\
        & \quad + \epsilon,
    \end{split}
\end{align}
where
\begin{equation}
f_{R_m}(R_m, R_s) = \frac{\partial f}{\partial R_m} = \frac{tR_s^2}{(1 - R_mR_s)^2},
\end{equation}

\begin{equation}
f_{R_s}(R_m, R_s) = \frac{\partial f}{\partial R_s} = \frac{t}{(1 - R_mR_s)^2}.
\end{equation}

When we convert the reflectance spectrum of foundation applied on top of skin to XYZ, the Z channel is:
\begin{align} \label{eq: km approximation}
    \begin{split}
        & Z = \int D_{65}(\lambda) \bar z(\lambda) (R_m + f(R_m, R_s))d\lambda \\
        & =\int \bar z'(\lambda) (R_m + f(R_m, R_s))d\lambda \\
        & = \int \bar z'(\lambda) R_m(\lambda)d\lambda + \int \bar z'(\lambda) f(R_m, R_s)d\lambda \\
        & \approx Z_m+f_{R_m}(r_m, r_s)\int \bar z'(\lambda)R_m(\lambda) d\lambda \\
        & \quad + f_{R_s}(r_m, r_s) \int \bar z'(\lambda) R_s(\lambda)d\lambda \\
        & \quad + \big (f(r_m, r_s) - r_m f_{R_m}(r_m, r_s) - r_sf_{R_s}(r_m, r_s) \big) \\
        & \quad \int \bar z' (\lambda) d\lambda \\
        & = Z_m + f_{R_m}(r_m, r_s)Z_m + f_{R_s}(r_m, r_s) Z_s \\
        & \quad + (f(r_m, r_s) - r_m f_{R_m}(r_m, r_s) - r_sf_{R_s}(r_m, r_s))C
    \end{split}
\end{align}


In \cref{eq: km approximation}, \(f_{R_s}(R_m, R_s)\) decreases as the thickness of foundation goes up. Therefore, when the thickness goes to infinity, \(f_{R_s}(R_m, R_s)\) goes to zero, meaning we will see less skin color as we apply more foundation on the face.
When the foundation thickness is zero, all the other terms in \cref{eq: km approximation} become zero and \(f_{R_s}(R_m, R_s)\) is one, then the blended color becomes the same as the original skin color. Additionally, both \(f_{R_s}(R_m, R_s)\) and \(f_{R_m}(R_m, R_s)\) increase as the transmittance goes up but \(f_{R_m}(R_m, R_s)\) increases slower than \(f_{R_s}(R_m, R_s)\), thus the skin color has a bigger impact on the resulting color when the foundation has a larger transmittance value.

To make linear Taylor expansion close to the original function, for each \(\lambda\), \(r_m\) needs to be close to \(R_m\) and \(r_s\) needs to be close to \(R_s\). Taylor expansion approximation equals the original function at \(R_m = r_m\) and \(R_s = r_s\). We want to find constants \(r_m\) and \(r_s\) such that
\begin{align*}
    \begin{split}
            & \quad f_{R_m}(r_m, r_s) \int \bar z'(\lambda) R_m(\lambda)d\lambda \\
            & = \int f_{R_m}(r_m(\lambda), r_s(\lambda))\bar z'(\lambda)R_m(\lambda)d\lambda , \\
            & \quad f_{R_s}(r_m, r_s) \int \bar z'(\lambda)R_s(\lambda)d\lambda  \\
            & = \int f_{R_s}(r_m(\lambda), r_s(\lambda))\bar z'(\lambda)R_s(\lambda)d\lambda 
    \end{split}
\end{align*}

We purposely make the three pieces of reflectance match the blue spike and green spike of the CIE XYZ standard observer color matching functions so that we can better approximate Kubelka-Munk model. We notice that $\bar z'(\lambda)$ is approximately symmetric between 400 and 500, and if $f_{R_m}(r_m, r_s)$ and $f_{R_s}(r_m, r_s)$ are linear, then the midpoint (e.g. $r_m = \frac{r_1 + r_2}{2}$) would be the optimal solution. However, it’s not true in our case. $f_{R_m}(r_m, r_s)$ is monotonously increasing on [0, 1] regarding $r_m$, and so as $f_{R_s}(r_m, r_s)$ regarding $r_s$. We still choose $r_m$ and $r_s$ to be the midpoint aiming to balance out the difference on the two sides, approximating the optimal result. 

Following the same method, we can compute for the Y channel too, but because X channel is widely spread across the whole spectrum and the shape is not symmetric, the approximation method might not work well.
Instead, we opt to separate the integration into 3 parts. For wavelength between 400 and 500, we apply similar method using midpoint to approximate the integration. Because the reflectance and transmittance is constant for wavelength between 600 and 700, we can easily calculate the integration over constant values. Finally we only need to integrate over the values for wavelength from 500 to 600.

\begin{figure*}[ht!]
    \centering
    \includegraphics[width=.32\linewidth]{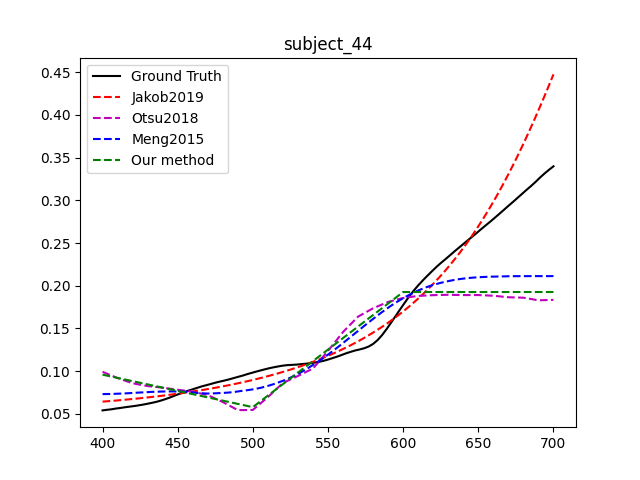}\hfill%
    \includegraphics[width=.32\linewidth]{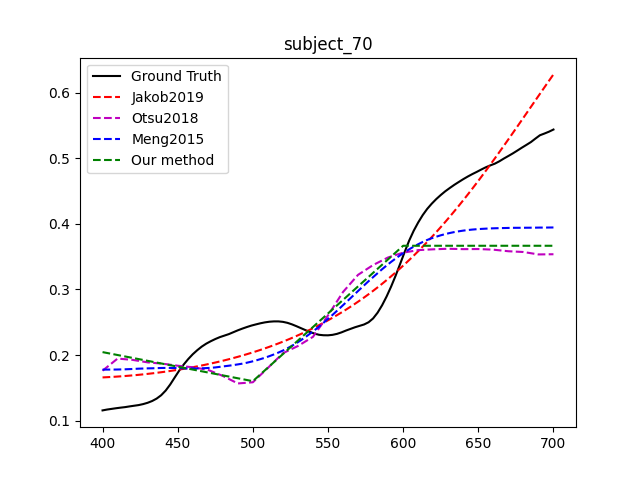}\hfill%
    \includegraphics[width=.32\linewidth]{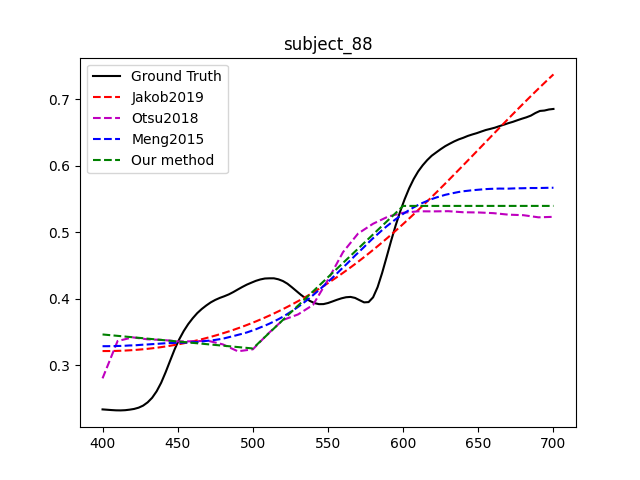}\\
    \caption{Visualization of skin reflectance comparison. Our method achieves similar performance with significantly less latency (See \cref{tab:relectance_computation_comparison})}.
    \label{fig:skin_reflectance_comparison}
\end{figure*}

\section{Results}
\subsection{Face semantic segmentation and skin reflectance estimation}
We used 3678 face images to train our semantic segmentation model, 967 images for validation and 252 images to evaluate different models performance. Please see \cref{fig:semantic-segmentation} and \cref{tab:face_segmentation} in supplementary materials for quantitative and qualitative comparison. 


To evaluate our skin reflectance estimation method, we use benchmark dataset\cite{skindata} containing 100 reference reflectance spectra of a wide range of human skin tones. \cref{fig:skin_reflectance_comparison} demonstrates a comparison of our method against existing reflectance estimation models across three distinct skin tones. Results demonstrate that our approach achieves comparable performance metrics while substantially reducing computational overhead (See \cref{tab:relectance_computation_comparison}).

\begin{table}[!ht]
    \centering
    \ra{1}
    \renewcommand\tabcolsep{4pt}
    \caption{Performance comparison for skin reflectance estimation methods of Jakob2019\cite{Jakob2019}, Otsu2018\cite{Otsu18}, Ment2015\cite{Meng2015} and our method. The reported latency is the computation time of reflectance spectrum given a single RGB (one pixel), while to process an image of the whole face, the latency difference will be multiplied by the size of the image.}
    \begin{tabularx}{0.47\textwidth}{lcccc}
        \toprule
        Method &  Jakob2019 & Otsu2018 & Meng2015 & Ours \\
        \midrule
        Latency (s) & 0.03 & 6e-4 & 0.94 & \textbf{3e-4}\\
        MSE  & 0.0016 & 0.0068 & 0.0039 & 0.0055\\
       \bottomrule
    \end{tabularx}
    \label{tab:relectance_computation_comparison}
\end{table}

\subsection{Virtual try-on synthesis with Kubelka-Munk theory}
\begin{wrapfigure}{l}{0.1\textwidth}
\centering
    \vspace{-8pt}
    \includegraphics[width=0.1\textwidth]{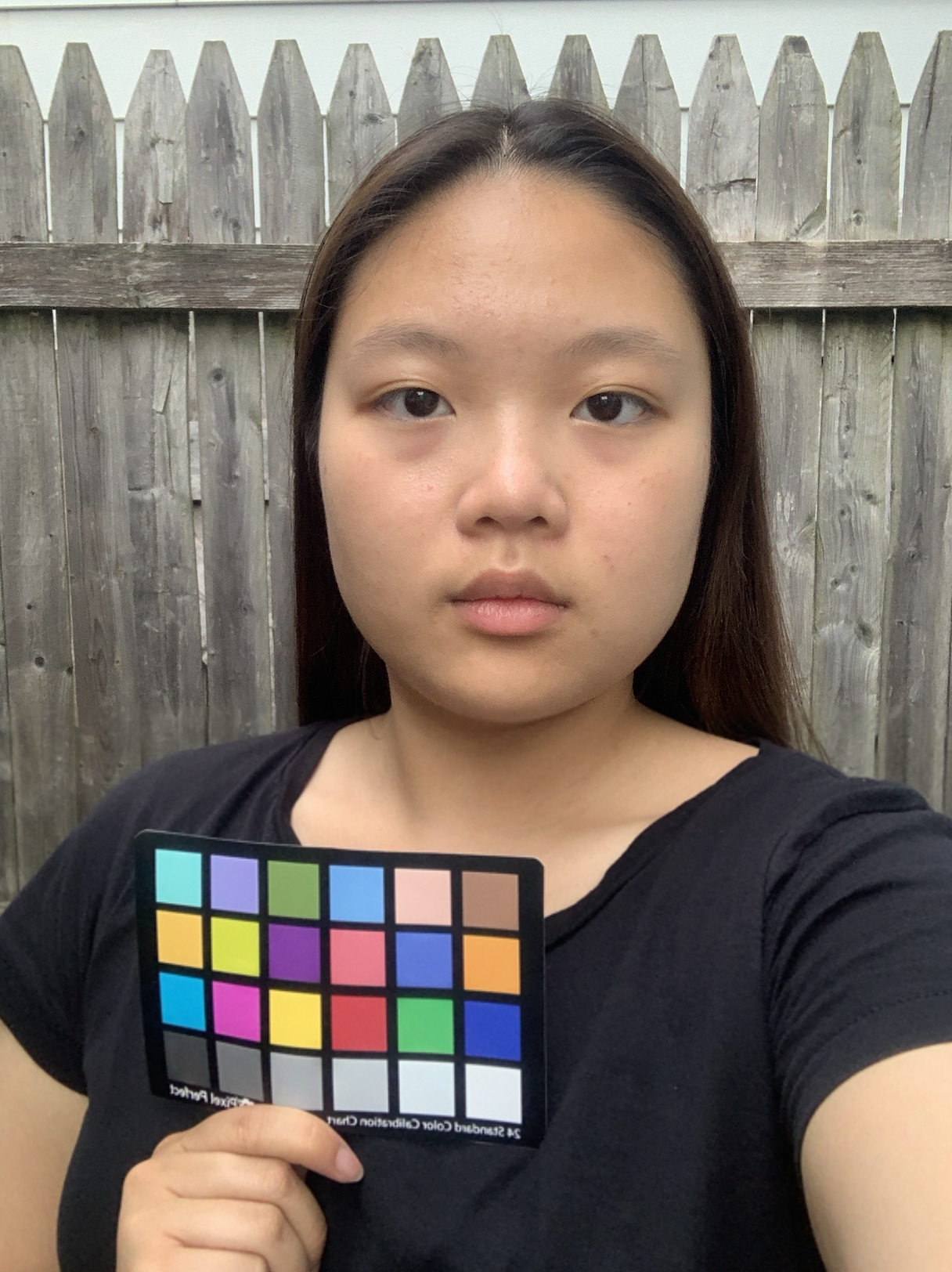}
\end{wrapfigure}

\textbf{Evaluating virtual try-on preformance:} To evaluate our methodology, we collected 10 people's selfie images with resolution 1000$\times$833 before and after applying foundation, along with the foundation product information such as color and coverage. We asked the participants to hold a color checker card when taking the picture, so that we were able to calibrate the color of images in different lighting conditions. Note that the intensity and the direction of the light might still differ in the ``before" and ``after" images, causing difference between the synthesized image and the real after-makeup image. 

We compare the latency of our approximation method for Kubleka-Munk model using Taylor expansion with vanilla integration over full visible spectrum running with Apple M2 Pro chip CPU in \cref{tab:computation_comparison}. \cref{tab:synthesis_similarity} demonstrates that the synthesized image using our approximation method is not visually distinguishable from the synthesized image with full visible spectrum integration.

\begin{table}[!ht]
    \centering
    \ra{1}
    \renewcommand\tabcolsep{4pt}
    \caption{Computation efficiency comparison between image synthesis with Kubelka-Munk model via full visible spectrum integration and our approximation method. We report the mean and standard deviation of the runtime in our experiments. Our method significantly improves the latency.}
    \begin{tabularx}{0.47\textwidth}{lcc}
        \toprule
        Method & Preprocessing (s) & KM Blending (s)\\
        \midrule
        Vanilla integration & 11.17 (0.07) & 0.925 (0.20)\\
        Our method & \textbf{6.85 (0.05)} & \textbf{0.17 (0.01)}\\
       \bottomrule
    \end{tabularx}
    \label{tab:computation_comparison}
\end{table}

\begin{table}[!ht]
    \centering
    \ra{1}
    \renewcommand\tabcolsep{3pt}
    \caption{Similarity between image synthesis with Kubelka-Munk model via full visible spectrum integration and our approximation method. We report the mean and standard deviation of SSIM, Delta E CIE2000 and LPIPS. Our approximation method is able to preserve the blending realism compared to integration over full visible spectrum.}
    \begin{tabularx}{0.47\textwidth}{lccc}
        \toprule
        Metric & SSIM & Delta E CIE2000 & LPIPS\\
        \midrule
        Value & 0.99 (6e-04) & 0.22 (0.076) & 8.5e-05 (3.3e-05)\\
       \bottomrule
    \end{tabularx}
    \label{tab:synthesis_similarity}
\end{table}

\begin{table}[ht]
    \centering
    \ra{1}
    \renewcommand\tabcolsep{3pt}
    \caption{Comparing synthesized images with real-world after-makeup images with LPIPS. Our method achieves the best LPIPS.}
    \begin{tabularx}{0.45\textwidth}{lcXX}
        \toprule
        \small Metric & \small Our result &  \small Vanilla \newline alpha blending & \small Alpha blending \newline with image decomposition \\
        \midrule
        \small LPIPS &  \textbf{0.226} &  0.236 & 0.231\\
       \bottomrule
    \end{tabularx}
    \label{tab:synthesis_similarity_quantitive_results}
\end{table}

\begin{figure*}[ht!]
\centering
    \includegraphics[width=0.9\linewidth]{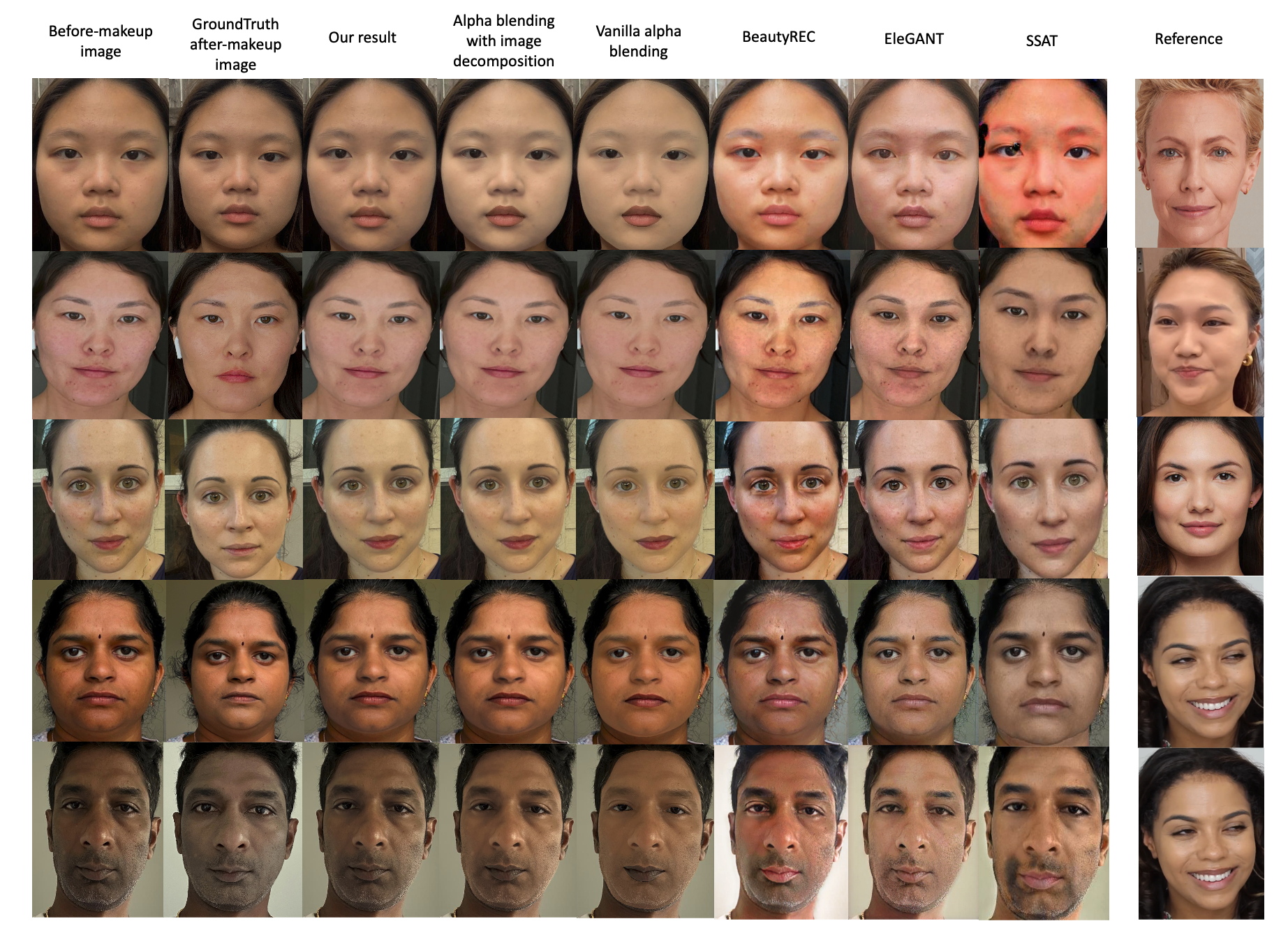}
    \caption{Real-world examples of comparison between our results with alpha blending using alpha=0.3, BeautyREC\cite{yan2022beautyrecrobustefficientcontentpreserving}, EleGANT\cite{yang2022elegant} and SSAT \cite{sun2021ssatsymmetricsemanticawaretransformer}. Alpha blending requires careful handpick for the alpha value to make the blending work well for different foundation products. Makeup transfer models fail to account for the interaction between skin tones and foundation shades and they are not robust to keep the facial features such as eyelashes, eyebrows, mustache, etc.}
    \label{fig:results}
\end{figure*}

\cref{tab:synthesis_similarity_quantitive_results} shows quantitative results comparing our model with alpha blending. Our method achieves a better LPIPS score comparing with real-world after-makeup image 100\% of the time in our experiment. \cref{fig:results} shows visual comparison results between our method, alpha blending and makeup transfer models. Makeup transfer models fail to maintain the lighting environment post foundation application and blend the foundation shade accurately with skin tone. We also find that for alpha blending an alpha value working well for some foundation product can be very off for other products, thus alpha blending requires careful handcrafting for alpha value to achieve good result for different foundation products. Additionally, alpha blending on the albedo layer deviates significantly from the real image, and applying alpha blending directly on the original image making makeup obscuring some facial details like eyebrows and nostrils. In contrast, our KM-based approach delivers the most color-accurate result while preserving all facial features, showcasing its effectiveness in achieving a realistic virtual makeup experience.

\section{Conclusion}

In this paper we presented an end-to-end framework designed to achieve realistic and scalable VTO experience for foundation makeup. Our framework leverages three key components: face semantic segmentation for precise makeup application, intrinsic image decomposition with an inpainting model for preserving light distribution and facial features, and an innovative approximation for Kubelka-Munk theory to achieve efficient and accurate color blending. Our approach doesn't rely on optical device measurements or auxiliary data, which makes it scalable to different foundation products and skin tones. Compared to traditional alpha-blending, our approach shows significant improvements in color accuracy and feature preservation. Future work will involve investigating methods to estimate the lighting environment within the image. This will further enhance color representation and realism. 



{
    \small
    \bibliographystyle{ieeenat_fullname}
    \bibliography{main}
}

\clearpage
\section*{\Large Supplementary Materials}

\subsection*{Face Semantic Segmentation}
\begin{figure}[!ht]
  \centering
  \includegraphics[width=0.9\linewidth]{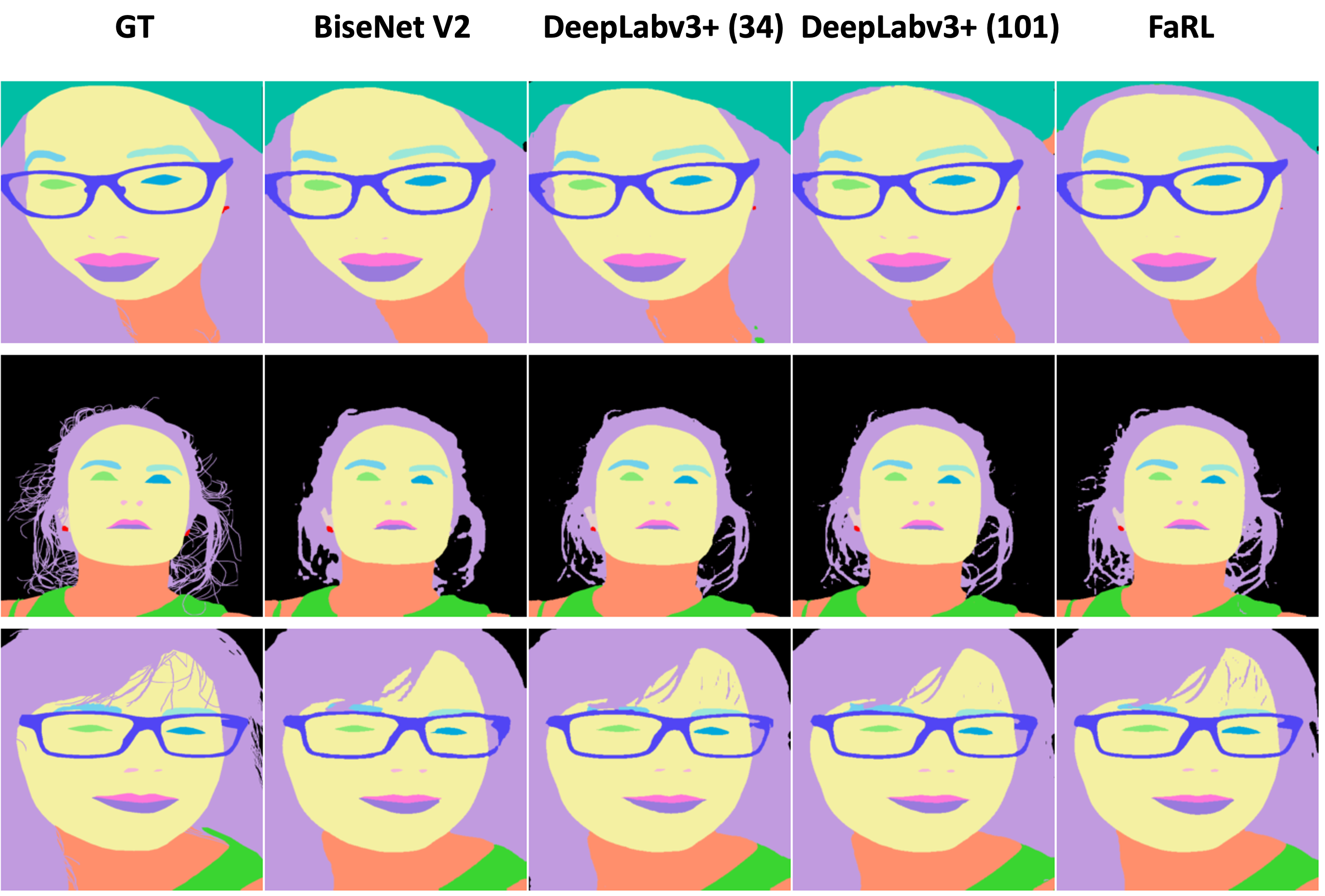}
  \caption{A comparison of different semantic segmentation models.}
  \label{fig:semantic-segmentation}
\end{figure}

We collected a dataset of 4897 face images for our semantic segmentation model (3678 images for training, 967 images for validation and 252 images to evaluation). We compared several baseline models for face semantic segmentation, including BiSeNetV2~\cite{yu2021bisenet}, U-Net~\cite{ronneberger2015u}, DeepLabv3~\cite{chen2017rethinking}, and FaRL~\cite{zheng2022general}. Table \ref{tab:face_segmentation} demonstrates the quantitative results for face, upper lips, lower lips and hair. Figure \ref{fig:semantic-segmentation} provides visual comparison of these models, from which FaRL is capable to capture more nuances in hair.

\begin{table*}[!b]
    \centering
    \ra{1}
    \renewcommand\tabcolsep{5pt}
    \caption{Face semantic segmentation comparison. We report Intersection-over-Union (IoU) for face, mouth, hair classes and mean value for all 17 classes.}
    \begin{tabularx}{1\textwidth}{lccccccc}
        \toprule
        &&& \multicolumn{4}{c}{IoU} \\
        Model & Pretraining Data & Model Size (M) & Face & Upper lips & Lower lips & Hair & Mean\\
        \midrule
        BiSeNetV2 & - & 4.3M & 0.95 & 0.82 & 0.85 & 0.81 & 0.68\\
        U-Net & - & 24.4M & 0.96 & 0.84 & 0.87 & 0.82 & 0.69\\
        DeepLabv3~(ResNet34) & -  & 22.4M & 0.96 & 0.83 & 0.86 & 0.82 & 0.69\\
        DeepLabv3~(ResNet101) & - & 45.7 M & 0.96& 0.83 & 0.86 & 0.82 & 0.69\\
        FaRL & LAION-Face 20M & 161.3M & 0.96& 0.81 & 0.85 & 0.84 & 0.72\\
       \bottomrule
    \end{tabularx}
    \label{tab:face_segmentation}
\end{table*}

\subsection*{Intrinsic image decomposition}
\begin{figure}[t!] 
    \centering
    \includegraphics[width=0.85\linewidth]{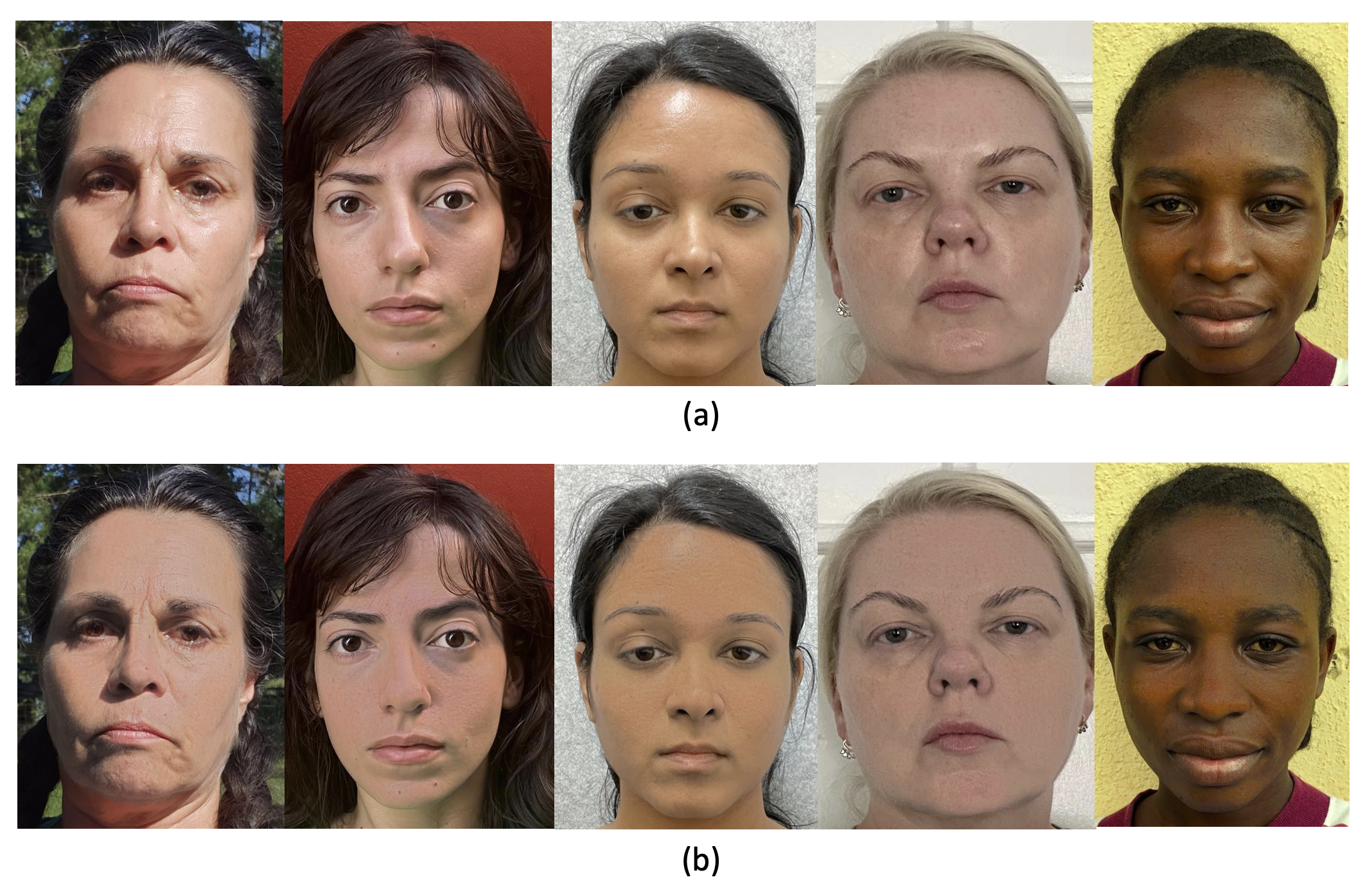}
    \caption{(a) Original images; (b) Synthesized images with highlight removed by inpainting model while preserving shadow.}
    \label{fig:inpainting}
\end{figure}

To extract the albedo color from facial image, we need to remove the shadow and highlight. However, when the highlights are too strong and appear as purely white pixels in the image, the information about the original albedo color becomes irretrievably lost. For that reason, we adopted LaMa inpainting model~\cite{suvorov2021resolution} to fill in the missing skin color based on the surrounding pixels not affected by highlights. Figure \ref{fig:inpainting} shows results after highlight removal while preserving the original shadow.

\end{document}